\def\year{2019}\relax
\documentclass[letterpaper]{article} %DO NOT CHANGE THIS
\usepackage{aaai19}  %Required
\usepackage{times}  %Required
\usepackage{helvet}  %Required
\usepackage{courier}  %Required
\usepackage{url}  %Required
\usepackage{graphicx}  %Required

\usepackage{tabularx}
\usepackage{hyperref}
\usepackage{booktabs}
\usepackage{makecell}
\usepackage{enumitem}
\usepackage{stfloats}
\usepackage{float}
\usepackage[hang,flushmargin]{footmisc}

\nocopyright

\begin{document}
\newcolumntype{L}[1]{>{\raggedright\let\newline\\\arraybackslash\hspace{0pt}}m{#1}}
\newcolumntype{C}[1]{>{\centering\let\newline\\\arraybackslash\hspace{0pt}}m{#1}}
\newcolumntype{R}[1]{>{\raggedleft\let\newline\\\arraybackslash\hspace{0pt}}m{#1}}

\title{The Toybox Dataset of Egocentric Visual Object Transformations}
% \author{AIVAS Lab\\
% Department of Electrical Engineering and Computer Science, Vanderbilt University\\
% PMB 351679, 2301 Vanderbilt Place, Nashville, TN 37235-1679, USA\\
% }
\author{
Xiaohan Wang\thanks{These authors contributed equally to this work.}
\qquad Tengyu Ma\footnotemark[1]
\qquad  James Ainooson
\qquad  Seunghwan Cha \\
\qquad \textbf{Xiaotian Wang}
\qquad \textbf{Azhar Molla}
\qquad \textbf{Maithilee Kunda}
\\
\normalfont {Department of Electrical Engineering and Computer Science, Vanderbilt University}\\
PMB 351679, 2301 Vanderbilt Place, Nashville, TN 37235-1679, USA\\
\texttt{\{xiaohan.wang, tengyu.ma, james.ainooson, seunghwan.cha,}\\
\texttt{xiaotian.wang, a.molla, mkunda\}@vanderbilt.edu} \\
}
\maketitle
\setcounter{footnote}{0}

%%%%%%%%%%%%%%%%%%%%%%%%%%%%%%%%%%%%%%%%%%%%%%%%%%%%%%%%%%%%%%%%%%%%%%%%%%%%%%%%%%%%
\begin{abstract}
  In object recognition research, many commonly used datasets (e.g., ImageNet and similar) contain relatively sparse distributions of object instances and views, e.g., one might see a thousand different pictures of a thousand different giraffes, mostly taken from a few conventionally photographed angles.  These distributional properties constrain the types of computational experiments that are able to be conducted with such datasets, and also do not reflect naturalistic patterns of embodied visual experience.  As a contribution to the small (but growing) number of multi-view object datasets that have been created to bridge this gap,  we introduce a new video dataset called Toybox that contains egocentric (i.e., first-person perspective) videos of common household objects and toys being manually manipulated to undergo structured transformations, such as rotation, translation, and zooming.  To illustrate potential uses of Toybox, we also present initial neural network experiments that examine 1) how training on different distributions of object instances and views affects recognition performance, and 2) how viewpoint-dependent object concepts are represented within the hidden layers of a trained network.
\end{abstract}

%%%%%%%%%%%%%%%%%%%%%%%%%%%%%%%%%%%%%%%%%%%%%%%%%%%%%%%%%%%%%%%%%%%%%%%
\section{Introduction}

Many recent breakthroughs in computer vision, such as for the problem of visual object recognition, have been driven by the creation and use of large-scale labeled image datasets collected from the Internet, with ImageNet being a canonical example \cite{deng2009imagenet}.  In a sense, these datasets have \textit{coevolved} with the algorithms that learn so successfully from them---researchers continue to develop algorithms that work better and better on the types of data distributions found on the Internet, and also continue to collect ever larger and more densely labeled datasets using similar online data collection methods.  

While the scientific advances and practical applications generated by these efforts have undoubtedly transformed the landscape of AI and machine learning, there are still many fundamental open research questions about how (and how well) intelligent agents can learn to recognize objects under very different types of training regimens.  In addition, there are many problems going beyond recognition that may require richer visual experiences with objects than are typically gathered online, for instance problems of visual commonsense reasoning or mental simulation.  Two areas that will especially benefit from studying such questions are:

\begin{figure}[ht]
    \centering
    \includegraphics[width=\linewidth]{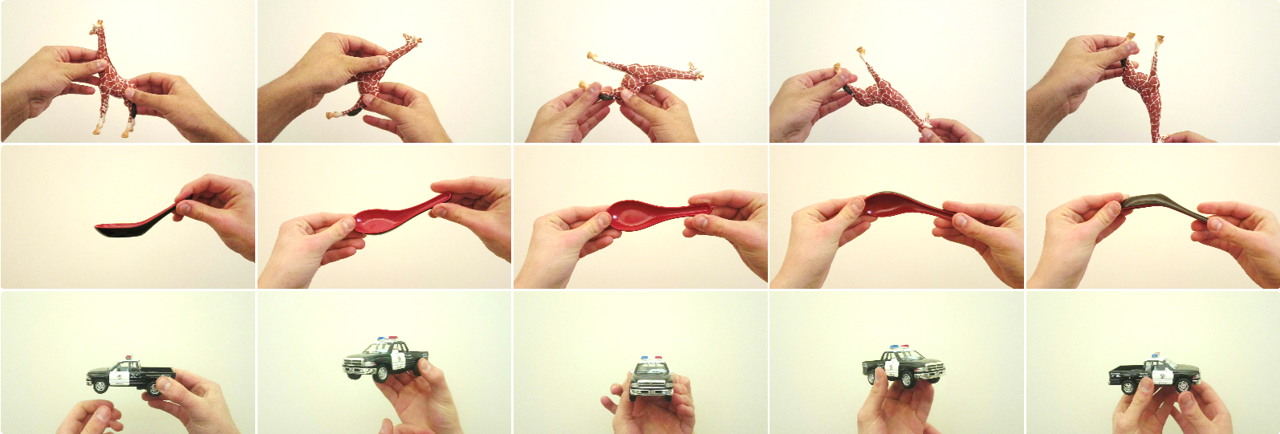}
    \caption{Toybox examples.  (Color adjusted for PDF view.)% The $1st$ row is a ryplus rotation sample from the animal super-category; The $2nd$ row is a rxplus rotation sample from the household super-category; The $3rd$ row is a rzplus rotation sample from the vehicle super-category.
    }
    \label{fig:toybox_example}
\end{figure}

\textbf{1. Research in AI + cognitive science to study the development of human object recognition.}  Research in developmental psychology has produced many interesting findings about the number and types of object instances children experience while learning category concepts.  In a fascinating diary study of her infant son, Mervis (\citeyear{mervis1987child}) observed that his initial concept of ``duck'' was likely based on seeing live ducks at a nearby park, his plush duck toy, a plastic duck rattle, a rubber duck, and a handful of other odd, duck-themed household objects, toys, and picture books, and was gradually generalized and pruned over time.  

Recent wearable-camera-based infant studies have found similar uneven distributions of experience across object instances, categories, and viewpoints, and these distributions also change with an infant's age \cite{james2014young,smith2018developing}.  It is likely not the case that infants learn \textit{despite} these irregularities, but rather that infant learning \textit{leverages} these distributional properties, e.g., through bootstrapping, curriculum learning, etc.   In other words, the infant's ``learning algorithm'' has coevolved alongside the neural, sensorimotor, cognitive, and sociocultural factors that combine to create an infant's visual world.  Thus, AI research that explores interactions between training distributions and learning algorithms is poised to play a critical role in the cognitive science of visual learning, including for understanding the effects of technology (e.g., advent of print media, television, and now Internet) on child development.

\textbf{2. Research in AI + robotics to advance the learning capabilities of embodied agents.}  Robots or other physically embodied agents (e.g., stationary cameras) may have access to online information but will likely also need to be able to learn from their immediate environment.  For example, a household robot may need to learn about specific objects in a person's kitchen through a series of naturalistic visuomotor interactions that generate complex, unevenly distributed, and heavily occluded object views.  What kinds of learning algorithms would be capable of learning from such data?  One-shot learning, active recognition, and various forms of active learning are all relevant to this question, and will benefit from the creation of richer visual datasets.

\textbf{The Toybox dataset.}  Here, we present a new dataset called Toybox (see example images in Figure \ref{fig:toybox_example}) designed to facilitate computational experiments on visual object recognition and related vision problems, especially in the context of studying aspects of embodied (e.g., human or robot) visual object experience, including: (1) \textit{continuously sampled views} of objects undergoing several different types of transformations, including rotation, translation, and zooming; (2) \textit{an egocentric perspective} (i.e., handheld, first-person views), which means that objects are held in naturalistic grips and thus are always partially occluded; (3) \textit{a range of everyday categories}, including household objects, animals, and vehicles; (4) \textit{a diversity of object instances}, with 30 distinct physical objects representing each category.

We also present two examples of the kinds of studies that Toybox is designed to support, including one experiment to studying the effects of instance and viewpoint diversity on recognition performance, and a second experiment to investigate how hidden layer neurons in a trained neural network respond to continuous variations in object pose.

%%%%%%%%%%%%%%%%%%%%%%%%%%%%%%%%%%%%%%%%%%%%%%%%%%%%%%%%%%%%%%%%%%%%%%%%%%%%%%%%%%%

\section{Multi-View Object Recognition Datasets}

As described above, ImageNet and many other widely used, ``Google Image Search''-type vision datasets contain only one image per real-world object \cite{deng2009imagenet}.  In addition, object viewpoints are constrained by the fact that most online images are created by adult humans using handheld camera devices \cite{torralba2011unbiased}.

\begin{figure}[b!]
    \centering
    \includegraphics[width=\linewidth]{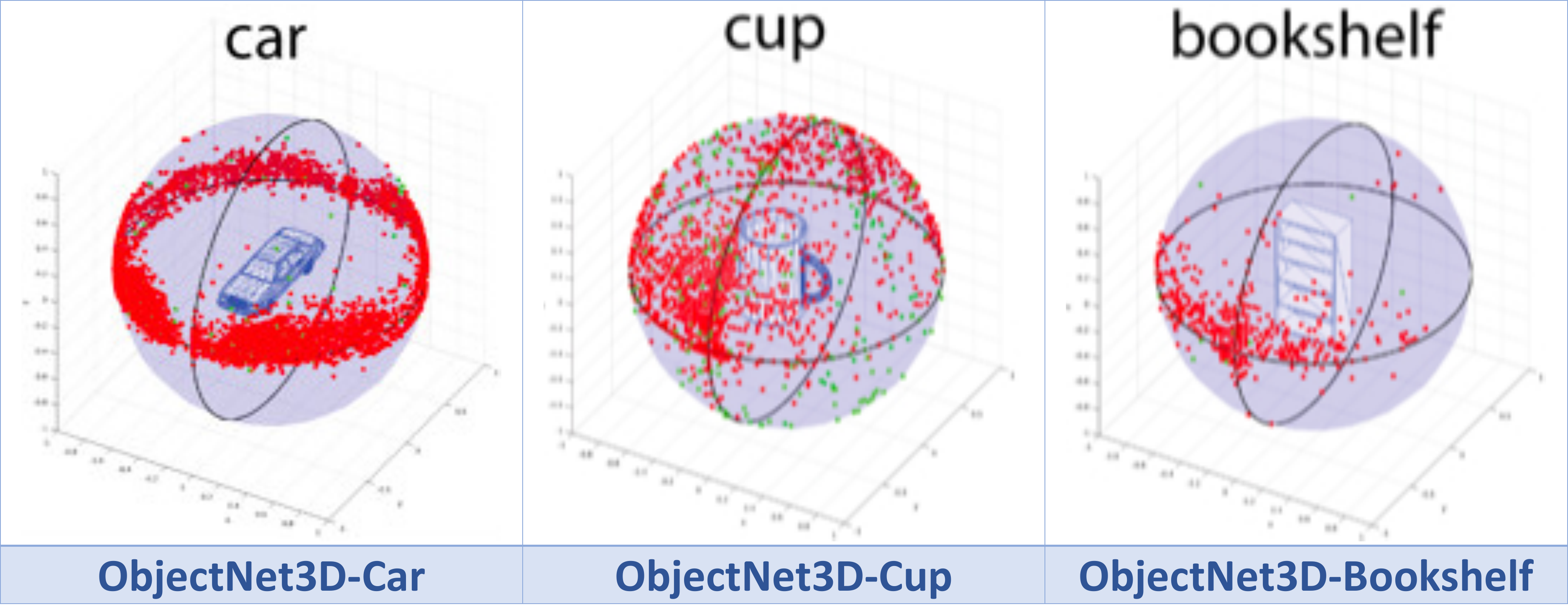}
    \caption{Viewpoint distributions of ImageNet categories, from the ObjectNet3D dataset \protect\cite{b:objectnet3d}.}
    \label{fig:objectnet3d_vp}
\end{figure}

Providing an interesting demonstration of this viewpoint bias, the ObjectNet3D dataset contains images from ImageNet annotated with the 3D pose of pictured objects \cite{b:objectnet3d}.  As shown in Figure \ref{fig:objectnet3d_vp}, different categories show very different viewpoint distributions, based on how people (adults) tend to encounter certain objects in everyday life.

%Many common object recognition datasets (e.g., ImageNet, Microsoft COCO, etc.) contain only one image per real-world object\cite{deng2009imagenet,lin2014microsoft}.  While these datasets have driven much exciting research in computer vision in recent years, they are, by their construction, limited in their applicability for supporting experiments to understand the training process of deep CNNs.

Taking a complementary approach, several vision datasets have been created to provide multiple views of the same physical object, as reviewed in Table \ref{table:datasets}.  These datasets are typically of two types: (1) \textit{discrete but structured object viewpoints} are collected with the help of a turntable, e.g., the NORB, RGB-D, and iLab-20M datasets \cite{lecun2004learning,lai2011large,borji2016ilab}; or (2) \textit{continuous but unstructured} objects viewpoints are collected using human handheld object and/or camera manipulations, e.g., the Intel Egocentric and CORe50 datasets \cite{ren2009egocentric,pmlr-v78-lomonaco17a}.

\begin{figure}[h!]
    \centering
    \includegraphics[width=\linewidth]{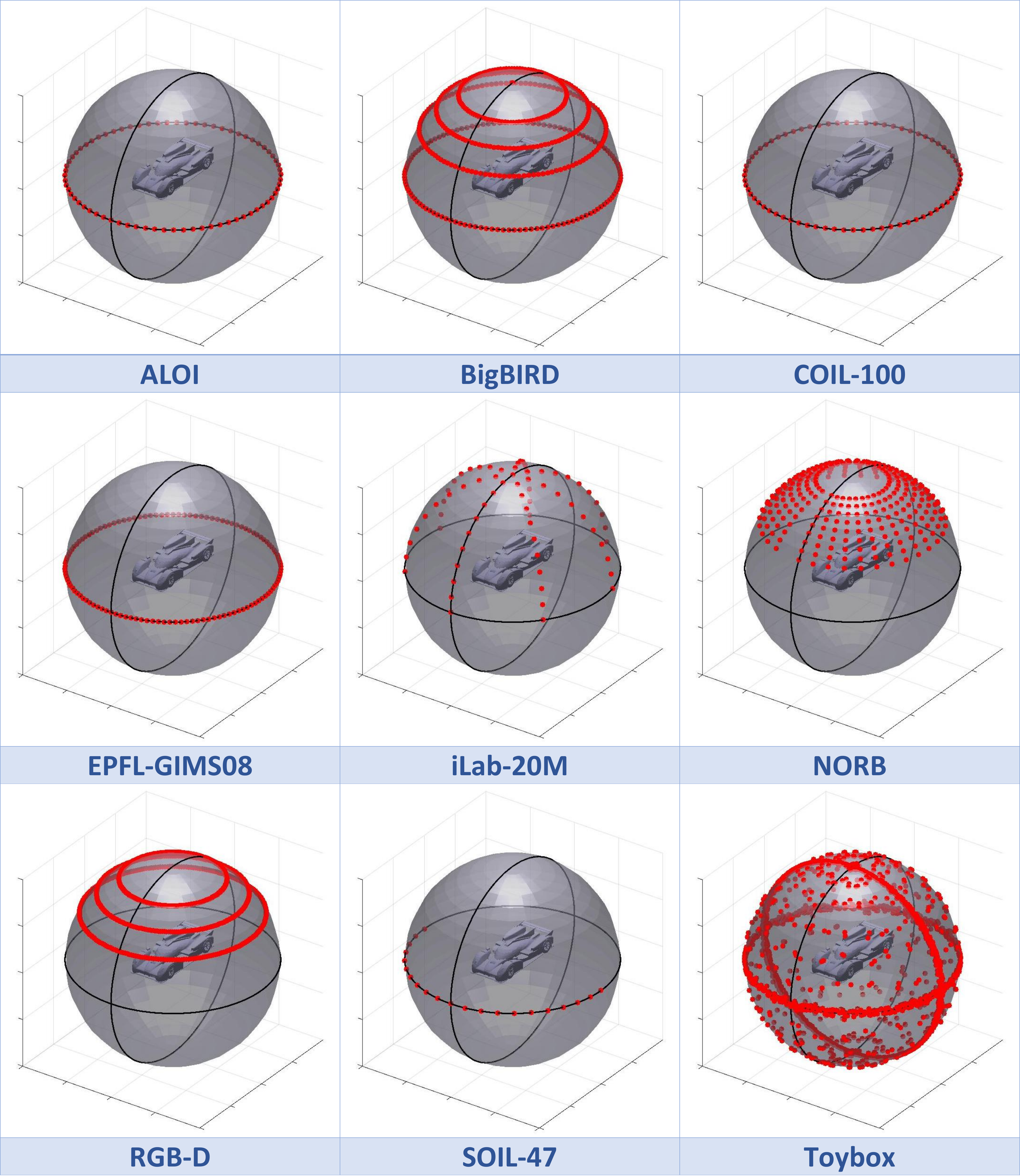}
    \caption{Comparison of viewpoint distributions across several multi-view datasets.  (Toybox off-axis viewpoints are estimations only and will vary from object to object.)}
    % \caption{The $1st$ row of view distributions are sampled from the ObjectNet3D paper \protect\cite{b:objectnet3d}. The $2nd-4th$ rows show the viewpoint distribution comparison among multi-view datasets.}
    \label{fig:viewpoints}
\end{figure}

We designed the Toybox dataset to capture the advantages of both approaches.  Toybox contains egocentric-perspective videos of the camera-wearer holding and manipulating various objects in structured ways, e.g. completing two full revolutions of an object along a specified axis of rotation at a (roughly) constant speed, among other types of transformations.  We also include a period of unstructured manipulation to capture a random assortment of off-axis views.  (Details about the dataset and collection methods are given below.)

Figure \ref{fig:viewpoints} shows the viewpoint distributions provided by several existing multi-view datasets, as well as by the Toybox dataset.  As can be seen in this figure, most of the existing multi-view datasets use turntables, and thus no bottom-facing views of objects are available.  Toybox aims to provide a more complete set of object views.  We do not know how valuable such views might (or might not) be, but at least having the data available will enable experiments to study viewpoint distributions and visual learning in more detail.

\begingroup
\begin{table*}
\setlength{\tabcolsep}{4pt} % Default value: 6pt
\fontsize{6.8pt}{10pt}\selectfont
\caption{Computer vision datasets that contain multiple real (i.e., not synthesized) images of the same physical object.}
\label{table:datasets}
\centering
\begin{tabularx}{\textwidth}
{@{} l C{3.8cm} c c c C{4.2cm} c r @{}}
\toprule
    % \multicolumn{2}{c}{Part} \\
    % \cmidrule(r){1-2}
    Dataset
    & Reference
    & Categories
    & Objs/Cat
    & Viewpoints/Obj
    & Other Variants
    & Imgs/Obj
    & Total Imgs\\
\midrule
    COIL-100
        & \cite{nene1996columbia} %(1996)
        & 100 % (household: mug, cup, can, etc.) %bubblegum, block house,
        & $\sim$1
        & 72 
        & n/a
        & 72
        & 7,200 \\
        
    SOIL-47
        & \cite{burianek2000soil} %(2000)
        & 47 % (household: lightbulb, mug, etc.) % crackers,  cereal,
        & $\sim$1
        & 21
        & lighting
        & 42
        & 1,974 \\
        
    NORB$^1$
        & \cite{lecun2004learning} %(2004) 
        & 5 % (human figure, car, truck, etc) %four-legged animal, airplane,  
        & 10
        & 648 
        & lighting
        & 3,888
        & 194,400 \\
        
    ALOI
        & \cite{geusebroek2005amsterdam} %(2005)
        & 1000 % (household: duck, tissues, etc.)% ball, pineapple, 
        & $\sim$1
        & 72
        & lighting
        & 111
        & 110,250 \\
        
    3D Objects on Turntable 
        & \cite{moreels2007evaluation}
        & 100
        & $\sim$1
        & 144
        & lighting
        & 432
        & 43,200 \\
        
    3D Object
        & \cite{savarese20073d} %(2007)
        & 8 % (household: bike, shoe, car, etc)%iron, mouse, cellphone, stapler, toaster
        & 10
        & 24
        & zooming
        & 72 
        & $\sim$7,000 \\
        
    Intel Egocentric$^{4,5}$
        & \cite{ren2009egocentric} %2009
        & 42 % (household: scissors, bowl, cup, wallet, etc.)
        & 1
        & \ \ various$^6$
        & background, manual activity
        & 1,600 
        & 70,000 \\
        
    EPFL-GIMS08
        & \cite{b:epfl-gism08}
        & 1 
        & 20
        & $\sim$120
        & n/a
        & $\sim$120
        & 2299 \\
        
    RGB-D$^{2}$
        & \cite{lai2011large} %(2011)
        & 51 % (household: bowl, stapler, etc.)%mushroom, keyboard, 
        & 3-14 
        & 750
        & camera resolution
        & 750
        & 250,000 \\
        
    BigBIRD$^{2}$ 
        & \cite{singh2014bigbird} %(2014)
        & 100 % (household: crayon, cereal, etc.)%toothpaste, 
        & $\sim$1
        & 600
        & n/a
        & 600 
        & 60,000 \\
        
    iCubWorld-Trfms.$^{3,4}$
        & \cite{pasquale2016object} %(2016)
        & 20 % (household: lotion, book, phone, etc.)% flower,
        & 10 
        & 150-200
        & lighting, background, zooming
        & $\sim$3,600 
        & $\sim$720,000 \\
        
    iLab-20M
        & \cite{borji2016ilab} %(2017)
        & 15 % (vehicles: boat, bus, car, tank, train, etc.)
        & 25-160 
        & 88 
        & lighting, background, focus
        & $>$18,480
        & 21,798,480 \\
        
    % Since Object3D is essentially a 3d annotation of ImageNet, it is hard to analyze the details inside 
    % ObjectNet3D
    %     & \cite{b:objectnet3d}
    %     & 100
    %     &   
    %     & various$^8$ 
    %     & n/a 
    %     & _ 
    %     & 90,127 \\
    
    CORe50$^{2,4,5}$ 
        & \cite{pmlr-v78-lomonaco17a} %(2017)
        & 10 %(household: plug, phone, scissors, etc.)
        % & lightbulb, can, glasses, ball, marker, cup, remote
        & 5 
        & \ \ various$^{6}$ % $\sim$1
        & indoor/outdoor, slight handheld movement 
        & $\sim$300
        & 164,866 \\
        
    eVDS 
        & \cite{e-VDS} 
        & 35 
        & 37-97 
        & \ \ various$^{6}$ 
        & n/a 
        & $>$144 
        & $\sim$420,000\vspace{3pt} \\ 
    
    \bfseries Toybox$^{4,5}$ 
        & \bfseries [this paper]
        & \bfseries 12 % (cup, mug, spoon, ball, cat, duck, horse, giraffe, car, truck, airplane, helicopter)
        & \bfseries 30
        & \bfseries $\sim$4,200
        & \bfseries translating, zooming, manual activity 
        & \bfseries $\sim$6,600 
        & \bfseries $\sim$2,300,000 \\
\bottomrule

\multicolumn{8}{@{} l}
{\textit{
$^1$ Stereo pairs not included in counts.\hspace{2pt}
$^2$ RGB-D video.\hspace{2pt}
$^3$ Updated counts from dataset website.
%}} \\
%\multicolumn{6}{@{} l}{\textit{
% $^4$ From arXiv preprint. \hspace{2pt}
$^4$ Handheld objects. \hspace{2pt}
$^5$ Egocentric video.
$^6$ Unstructured viewpoint distributions.
}} \\

\end{tabularx}
\end{table*}
\endgroup

\section{Toybox: Dataset Organization and Collection}
\label{sec: dataset}
% \begin{figure}[t]
%     \centering
%     \includegraphics[width=\linewidth]{1_images_working/methodology_3.jpg}
%     \caption{Recording Process}
%     \label{fig:recording_process}
% \end{figure}

\begin{figure*}[t]
  \centering
  \includegraphics[width=\linewidth]{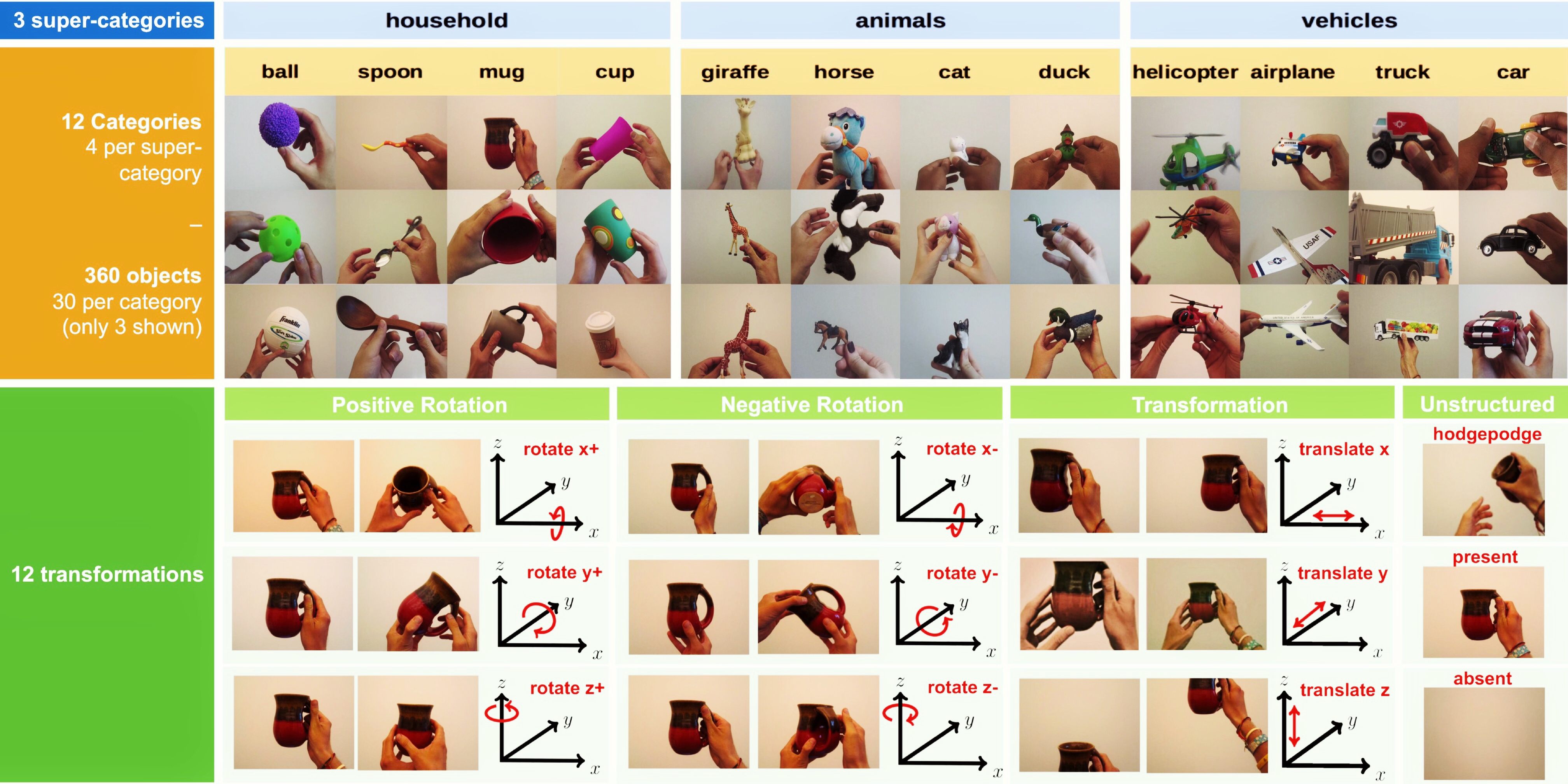}
  \caption{Toybox overview. Toybox contains 12 categories with 30 individual physical objects per category.  There are 12 video clips per object.  Each clip contains a defined transformation of the object: two full revolutions for rotation clips and three back-and-forth shifts for translation clips.  A final ``hodgepodge'' clip contains unstructured object motion, mostly rotations.  Please see Supplementary Video 1 for representative clips.}
  \label{fig:toybox_overview}
\end{figure*}

Figure \ref{fig:toybox_overview} provides an overview of the Toybox dataset.  Representative video clips from Toybox can be viewed in Supplementary Video 1.  This section provides details about the design of the dataset and our recording methods.

\textbf{Selection of categories and objects.}  Toybox contains 12 categories, roughly grouped into three super-categories: \textit{household items} (cup, mug, spoon, ball), \textit{animals} (duck, cat, horse, giraffe), and \textit{vehicles} (car, truck, airplane, helicopter).

To maximize the usefulness of Toybox for comparisons with studies of human learning, all 12 of these categories are among the most common early-learned nouns for typically developing children in the U.S. \cite{fenson2007macarthur}.  Categories were also selected to provide shape variety in each super-category (e.g., spoon vs. ball, duck vs. cat, etc.) as well as shape similarity (e.g., cup vs. mug, car vs. truck, etc).

Each category contains 30 individual physical objects. For both animals and vehicles, we cannot include real objects, and so objects are either realistic, scaled-down models or ``cartoony'' toy objects.  
Objects were purchased mostly in local stores, with some acquired online.  Individual objects were selected to provide a variety of shapes, colors, sizes, etc., and can be considered a representative sampling of typical objects available in the U.S.
% (region omitted for blind review).% (see Figure \ref{fig:toybox_overview} and \ref{fig:size-distribution}).

\textbf{Recording devices.}  Videos were recorded using Pivothead Original Series wearable cameras, which are worn like sunglasses and have the camera located just above the bridge of the wearer's nose.  Camera settings included:  video resolution set to \textit{1920} x\textit{1080}; frame rate set to \textit{30 fps}; quality set to \textit{SFine}; focus set to \textit{auto}; and exposure set to \textit{auto}. 

\textbf{Canonical views.}  For each category, we defined a canonical view of the object, roughly centered in front of the camera-wearer’s eyes.  For example, mugs start in an upright position with handle to the right.  Animals and vehicles start in an upright position facing towards the left.  %For cups, the canonical view is defined as the object held upright.   For mugs, the canonical view is defined as upright with the handle pointing to the right.  For spoons, likewise, the canonical view has the handle pointed to the right and the bowl of the spoon turned up.  For animals and vehicles, the canonical view is defined as the object facing towards the left (or standing with its head towards the left side, if its face is not aligned with its body).

\textbf{Video clips.}  For each object, a set of 12 videos was recorded, as shown in Figure \ref{fig:toybox_overview} and Supplementary Video 1.  Each clip is $\sim$20 seconds long, with the exception of absent/present video clips, which are $\sim$2 seconds long.  For rotations, each clip contains two full revolutions of the object; for translations, each video contains three back-and-forth translations starting from the minus end of each axis. Rotations and translations were controlled to have an approximately constant velocity over the 20-second duration of the video.  %To do this, we developed a set of audio ``temporal instruction templates'' that camera-wearers would listen to while creating each video.  
Thus, the pose of the object in every frame of a given video clip can be estimated according to its time.

\textbf{Recording procedures.}  Objects were semi-randomly assigned to individual camera-wearers (members of our research lab) such that no individual was over-represented in any category or object size class, to reduce any biases related to specific personal attributes or individual hand gestures.  All videos were collected in an indoor setting against an off-white wall.  Recordings were made across various times of day and lighting conditions, and so there is variation in lighting across different objects (as can be seen in Figure \ref{fig:toybox_overview}).

% \begin{figure*}[t]
%     \centering
%     \includegraphics[width=\linewidth]{count_cluster_v2}
%     \caption{Object size distribution of our Toybox dataset. The size of each object was estimated by the total pixels from a cropped image with a minimal size that contains the object from the "present" video clip.}
%     \label{fig:size-distribution}
% \end{figure*}

%\textbf{Video-to-image conversion for experiments.}  Object videos were converted to images in jpeg format using FFmpeg. Due to the limited field of view of the Pivothead wearable camera, we found that in some images, the object was almost or completely out of the image frame. To eliminate these ``blank'' images, the whole dataset was first used to re-train the Inception v3 neural network (as described in Section \ref{sec:methods}), with all 12 categories plus a 13th ``blank'' category that contained images from the ``absent'' videos, i.e., videos that recorded background only (see Table \ref{table:manipulations}).  The re-trained neural network was then applied back to the whole dataset to screen for ``blank'' images.  About 10,000 ``blank'' images ($\sim$1\%) were found using this classifier and, after manual confirmation, were subsequently removed for the experiments that are described next.

%%%%%%%%%%%%%%%%%%%%%%%%%%%%%%%%%%%%%%%%%%
\section{Sample Experiments Using Toybox}
\label{sec:methods}

For initial, proof-of-concept object recognition experiments with Toybox, we use the transfer learning methodology appearing in many recent studies, e.g., \cite{bambach2016active,pasquale2016object}, which involves re-training the last layer of a pre-trained, deep convolutional neural network.  %We used the ImageNet ILSVRC 2012 pre-trained Inception v3 network as a fixed feature extractor, and then re-trained the last layer as specified for each experiment. 

In particular, we use the Inception v3 network, as implemented in the Tensorflow software library \cite{tensorflow2015-whitepaper}.  
Inception is a representative convolutional neural network that has been shown to be highly successful in recognition tasks \cite{szegedy2016rethinking}.  The Inception v3 model we used here was pre-trained on the ImageNet ILSVRC 2012 dataset, which contains 1.2 million images from 1,000 categories.
More than half of the Toybox categories do appear in the original 1,000 categories used for pre-training---except for helicopter, giraffe, horse, and duck.  Our ongoing research includes training from scratch (see future work). %Notice that although the ILSVRC 2012 dataset did not contain the full Toybox dataset, the entire ImageNet dataset did.  

Experiment 1 studied instance diversity and view diversity, and Experiment 2 studied viewpoint-dependent hidden layer representations, as described in more detail below.

%------------------------------------------------------------------------

%%%%%%%%%%%%%%%%%%%%%%%%%%%%%%%%%%%%%%%%%%%
\subsection{Experiment 1: Using Toybox to study the effects of instance diversity and view diversity on recognition}
\label{subsec:retrain}

In Experiment 1, we re-trained the last layer of the ILSVRC 2012 pre-trained Inception v3 network using images from Toybox, and then tested recognition performance using images from the same categories from ImageNet.  Test performance is measured as the top-1 error rate.

Note that the choice of using ImageNet images (instead of hold-out Toybox images) for testing was deliberate.  We aimed to explore how well training on a small number of handheld, often toy objects would be able to generalize to the very different objects/views in ImageNet (e.g., training on toy cats to recognize real cats). Certainly other testing approaches would also be interesting and will be pursued in future work.  We constructed this ImageNet test set to contain 100 images/category across the 12 Toybox categories.

%\textbf{Test set.}  In Experiment 1, we first downloaded images from ImageNet from the same 12 corresponding categories to our Toybox dataset. We then sampled 1100 images/category for training and 100 images/category for testing. 

\textbf{Instance diversity.}  We first looked at the effect of \textit{instance diversity} on recognition performance by varying the number of individual physical objects per category in the training dataset, while keeping the total number of training images per category fixed at 1100 and uniformly drawn from the various video clips contained in the Toybox dataset. 

For example, with one object per category, each of the 12 categories is represented by 1100 images of a single object from that category.  With two objects per category, each category is represented by 1100 images uniformly drawn across two objects (550 images per object on average). %As a comparison, we also used images from ImageNet for retraining, with 1100 images per category (which corresponds to having 1100 objects per category, since ImageNet essentially has one image per physical object).

\begin{figure}[t!]
    \centering
    \small
    \includegraphics[width=\linewidth]{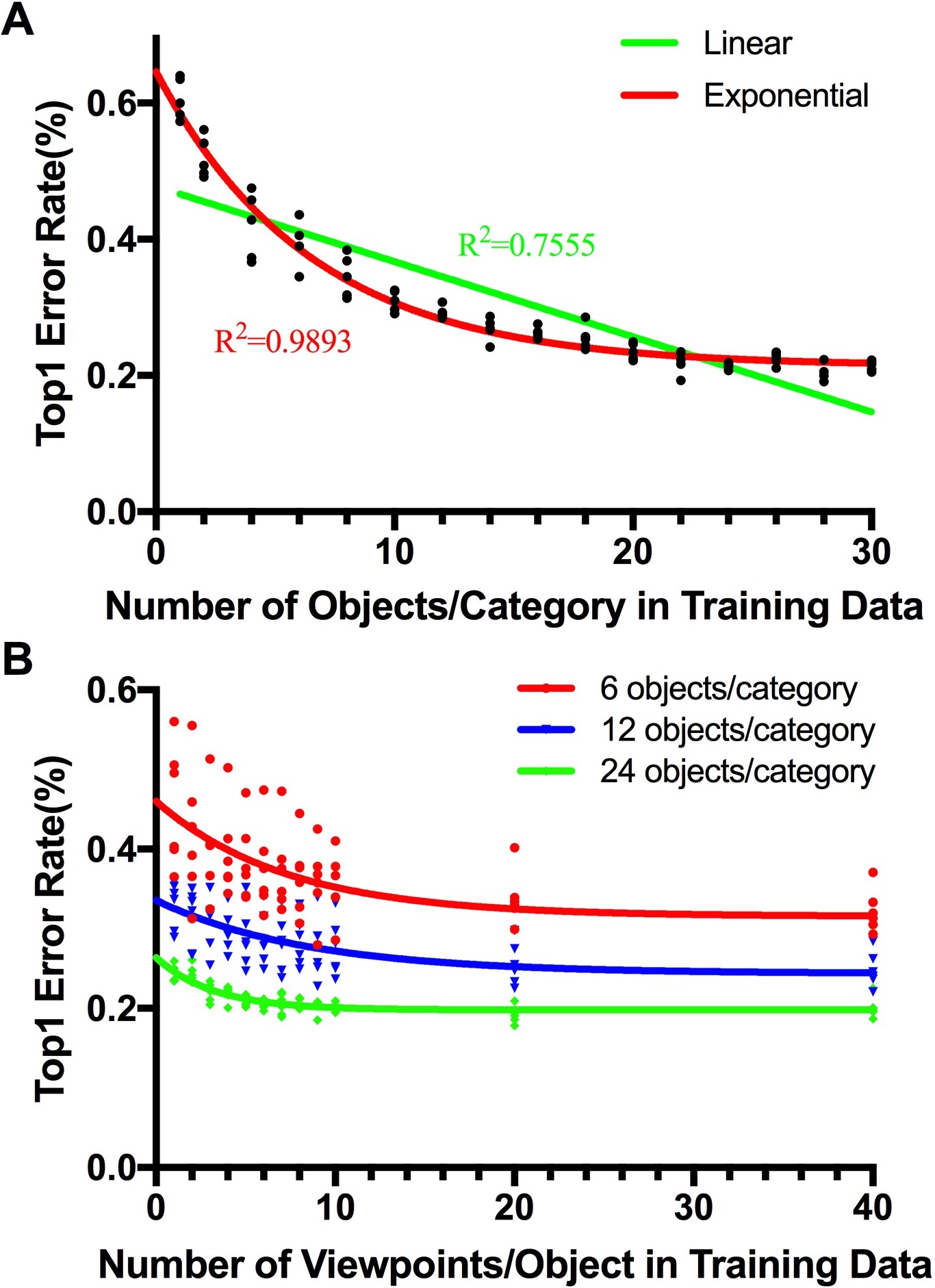}
    \caption{Effects of object diversity and view diversity on recognition performance, measured as top-1 error rate on an ImageNet-sourced test set.  Each trial was run 5-6 times, shown as individual data points. \textbf{A}. Recognition as a function of instance diversity, i.e., number of objects per category in the Toybox training set, with the total size of the training set held fixed. \textbf{B}. Recognition as a function of view diversity, i.e., number of images per object in the Toybox training set, ranging from 2 to 40 images per object. %The effect of view diversity was also tested with different object numbers (i.e., 6, 12, 24) per category. For instance, the total number of training images per category varies from 24 to 480 for the 12 objects per category group. All data points were from 5-6 independent experiments with different objects selected randomly.
    }
    \label{fig:object-effects}
\end{figure}

Results from this experiment are shown in Figure \ref{fig:object-effects}A.  
A training set with images of only a single Toybox object per category (i.e., 1100 images per object) yields an average error rate of 60.63\%, which while not excellent, is well below the random-guessing baseline error rate of 91.7\%. Adding a second object (i.e., about 550 images per each of two objects) further reduces error to 51.98\%. Adding more objects per category (with total training images per category fixed at 1100) continues to improve performance significantly, with our final experiment using 30 objects per category yielding an average error rate of 21.43\%.

We also characterized the performance improvement by computing best-fit lines using both linear and exponential models. As shown in Figure \ref{fig:object-effects}A, the exponential curve yields a better fit. Therefore, at least from the perspective of this model fitting, it appears that increasing object diversity will reduce the error rate in an exponential manner, with much greater improvements in performance for the first few added objects, and smaller increases thereafter (especially after about 20 individual objects).

\textbf{View diversity.}
We also looked at \textit{view diversity}, by varying the number of images per object included in the Toybox training set.  By sampling these images uniformly across all Toybox video clips, the number of images per object can be used as a proxy for views per object.  We conducted this experiment under three conditions, with the total number of objects per category fixed at 6, 12, and 24, respectively. 

For example, for 12 objects per category condition, we varied the total number of images per object from 2 to 100, drawn uniformly across all 12 objects.  Specifically, if we pick 2 images per object, the training dataset would have $2\times12=24$ images per category, and similarly, if we pick 100 images per object, the training dataset would have $100\times12=1200$ images per category.

Figure \ref{fig:object-effects}B shows results from this experiment.  (Although we experimented with numbers of images per object up to 100, we noticed a near constant error rate once this number exceeded 40, and so the graph in the figure is truncated at $x=40$.) %Due to this observation, we truncated the x-axis at 40 in \ref{fig:object-effects}B to give a better illustration. 
Results across the three conditions show similar trends, and so we focus our discussion here on the 12 objects/category condition (blue data points and curve).  % Without loss of generality, concentrating on the blue points and the exponential fitted curve (12 objects per category), 

With a single image per object, the average top1 error rate is 33.0\%. This error rate is subsequently reduced to 27.5\% if we have 10 images per object, and is further reduced to 25.6\% and 24.8\% for 20 and 40 images per object, respectively.  As with object diversity, the effects of view diversity appear to show an exponential trend, with only modest improvements after about 5-10 views per object.  %Increasing the number of views per object can apparently improve the performance of the classifier at the very beginning, i.e., 40 images per object achieve more than 8\% lower error rate than the 1 image per object. On the contrary, if we keep increasing the number of images, for example, 100 images per object with average error rate 23.9\%, the improvement becomes limited with only a 1.1\% error rate decrease compared to the result obtained with 40 images per object.

\begin{figure*}[!ht]
    \centering
    \includegraphics[width=\linewidth]{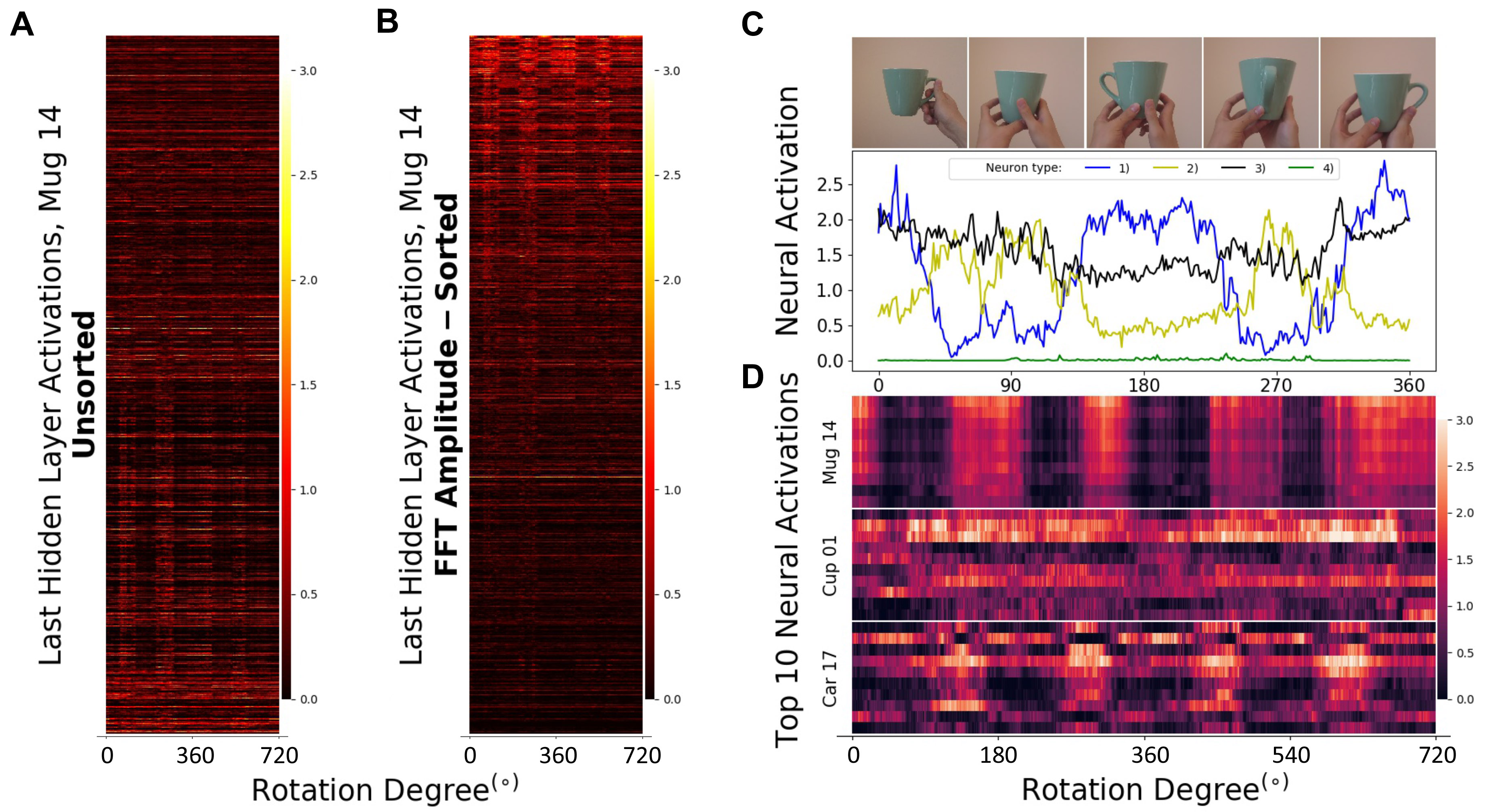}
    \caption{A novel method to identify neurons that correlate with a rotating object using our Toybox video dataset. 
    \textbf{A}. Temporal raster plot of neural activations in the final hidden layer of the pre-trained Inception v3 network while ``watching'' a rotating mug.  Each row shows an individual neuron, and the x-axis depicts time/rotation.  %The 2048 bottleneck neurons are lined up vertically, and each horizontal line represents the activation of a specific neuron over the two cycles of mug rotation. 
    \textbf{B}. The same neurons sorted based on their FFT amplitude from high (top) to low (bottom). Note the stripe pattern on the top half of the plot showing strong periodicity. 
    \textbf{C}. Four different types of neurons identified using this method. 
    \textbf{D}. Comparison of FFT analysis of mug, cup, and car, showing top 10 neurons after FFT amplitude sorting.}
    \label{fig:FFT_analysis}
\end{figure*}

%-------------------------------------------------------------------------
%%%%%%%%%%%%%%%%%%%%%%%%%%%%%%%%%%%%%%%%%%%
\subsection{Experiment 2: Using Toybox to study viewpoint-
dependent hidden layer representations}
\label{subsec:representation}

One fundamental question related to deep neural networks and object recognition is how objects are represented within the hidden layers of a network. We propose that the structured transformations of objects in Toybox videos may help us to better understand these representations.

Similar to Experiment 1, we used the ILSVRC 2012 pre-trained Inception v3 network, with the final output layer retrained on Toybox data with 1100 images per category selected across all Toybox objects.  We also used the same ImageNet-sourced test set with 100 images per category.

\textbf{Quantifying neuron temporal activation profiles.}
We began by studying the ``temporal'' activation profiles of neurons in the last hidden layer of the Inception network, while the network is receiving a sequence of Toybox input images depicting a mug rotating along the z(+) axis for two full cycles.  
Figure \ref{fig:FFT_analysis} shows visualizations of these activations.  

In particular, Figure \ref{fig:FFT_analysis}A depicts the activations over time of all 2048 neurons in the final hidden layer of the network.  Each row shows the activations of an individual neuron (unsorted in this subfigure), and the x-axis indicates time, which also approximates the rotation degree of the mug.  (This visualization method is adapted from the temporal raster plots used in neural physiology research.)
%To do this, we borrowed the "temporal raster plot" from the neural physiology research and plotted the "neural activation" of each neuron in the bottleneck layer. 
The various neuron ``firing patterns'' are clearly heterogeneous: some neurons are constantly firing throughout the two rotation cycles, some remain silenced, and some fluctuate as the mug rotates.

To differentiate these neuron types, we applied a Fast Fourier Transformation (FFT) to the activation of each neuron over the two rotation cycles.  %Because horizontal flipping is included in the data augmentation from initial training of Inception v3, we reasoned that the pre-trained network should not be able to differentiate horizontally symmetric images. Therefore 
To capture general viewpoint trends, we focused our FFT analysis on a frequency of 4 (i.e., four cycles within the 20-second long video that contains two complete rotations).  We then sorted the 2048 neurons shown in Figure \ref{fig:FFT_analysis}A based on their FFT amplitude at the frequency of 4---larger amplitudes indicate more robust oscillations.   Figure \ref{fig:FFT_analysis}B shows a visualization of the same neurons but sorted (top-to-bottom) by their FFT amplitudes.

Since FFT analysis also returns the phase information (positive phase correlates with handle presence, negative phase correlates with handle absent in this case), we were able to identify four different types of neurons based on their activation profiles.  Examples of these four types are shown in Figure \ref{fig:FFT_analysis}C and also in Supplementary Video 2: (1) neurons that fire when the mug handle is present (blue line); (2) neurons that fire when the mug handle is \textit{behind} or \textit{in front of} the mug body (yellow line); (3) neurons that fire throughout the video clip (black line); and (4) neurons that do not fire at all (green line, these neurons presumably do not contribute to the representation of the mug).

We also tested this FFT analysis method on objects from other categories using our Toybox videos. As shown in Figure \ref{fig:FFT_analysis}D, the ability to identify robust oscillating neurons mainly depends on the degree of asymmetry of the object along the z-axis. For instance, we were able to identify neurons with more robust oscillation for a mug and a car than for a cup (which is symmetric along the z-axis).

\begin{figure*}[!ht]
    \centering
    \includegraphics[width=\linewidth]{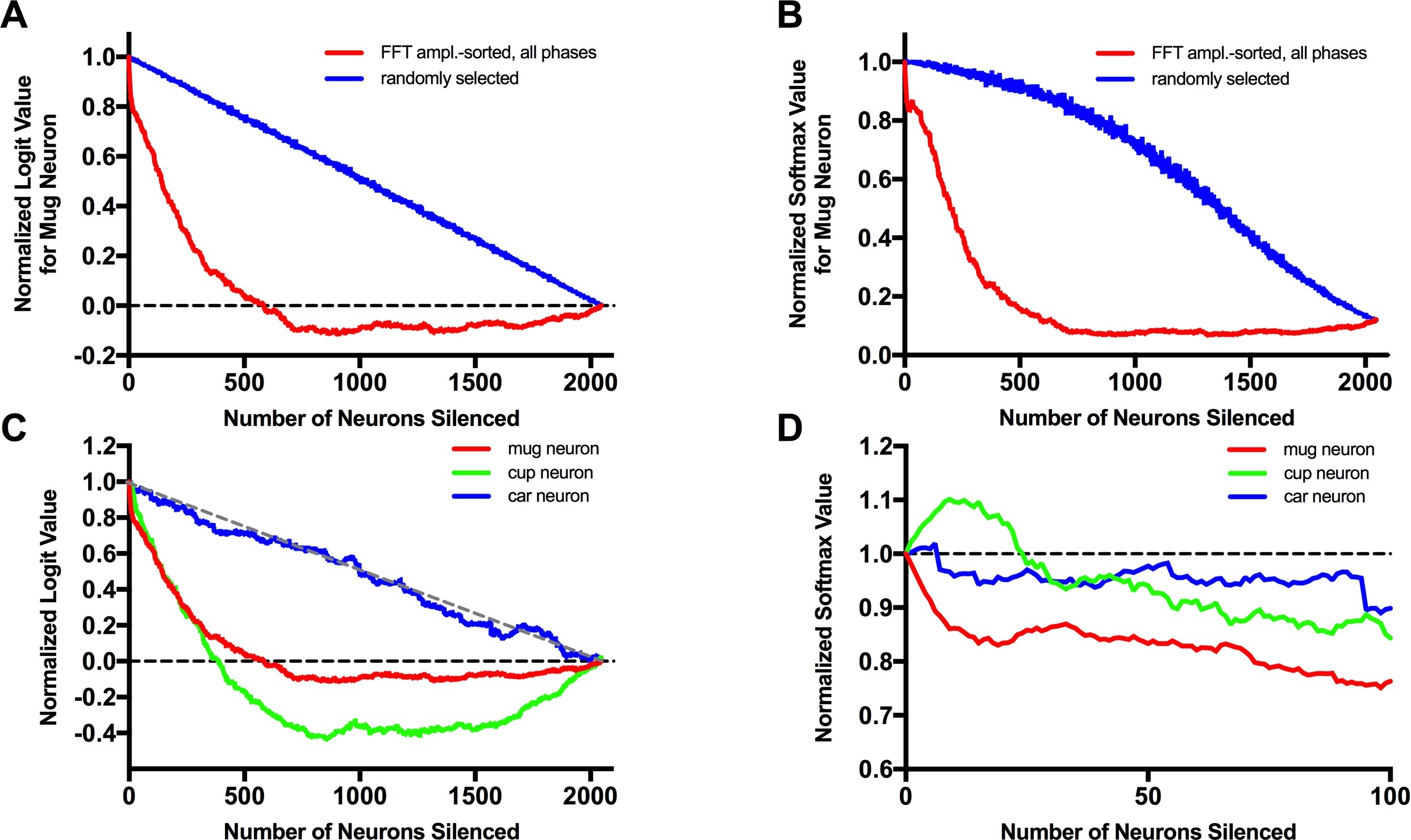}
    \caption{Effects of silencing hidden layer neurons on output layer activations, averaged over 100 images per category from ImageNet-sourced test set. \textbf{A}. Silencing top N neurons based on FFT amplitude sorting leads to a much steeper reduction in normalized logit value of the mug output neuron. The blue line shows the reduction rate of silencing N randomly selected neurons as a control. \textbf{B}. Similar to A, showing softmax values instead of logit values. \textbf{C}. Silencing \textit{mug-preferred neurons} (MPNs) has a similar effect on the logit value of the cup output neuron but has no effect on that of the car output neuron. \textbf{D}. Zoomed-in plot of softmax values, showing that silencing the top $\sim$20 neurons decreases mug prediction confidence while increasing cup prediction confidence, consistent with the fact that the majority of these neurons correlate with the presence of the mug handle. %A total of 100 images from the corresponding category of the test set were fed to the network to calculate the average of the normalized logit value of the given output neuron.
    }
    \label{fig:neuron_silencing}
\end{figure*}

\textbf{Effects of neuron silencing.}
To investigate the implicit representations of the various types of neurons identified above, we performed a neuron silencing/lesion experiment by selectively ``zeroing out'' the activations of certain neurons in the last hidden layer, and then observing effects on recognition performance.  %We then ran the retrained Inception network on each of the mug images from the test set and asked what is the impact if we silence a certain number of neurons from the bottleneck layer. 
Figure \ref{fig:neuron_silencing} shows results from these experiments, where the logit and softmax values for particular output neurons are shown as averages over 100 images from various categories in the ImageNet-based test set.

First, for test images from the mug category, we silenced N neurons (N varying from 0 to 2048) in the last hidden layer and examined the changes of both logit and softmax values of the mug neuron in the output layer.  As shown in Figure \ref{fig:neuron_silencing}A, silencing 0 of these hidden layer neurons has no effect, while silencing all 2048 neurons reduces the normalized logit value of the mug neuron to 0. Randomly silencing a subset of N neurons leads to a linear reduction of the logit value with respect to N. However, if we silence neurons based on the FFT amplitude sorting as shown in Figure \ref{fig:FFT_analysis}B (i.e., first silencing the neuron with the highest FFT amplitude, then the top two, then top three, and so on), we observed a much steeper drop in mug logit value at the beginning. After $\sim$700 neurons, silencing has no more reduction effect on the mug logit value. Similar effects can be seen with the softmax value of the mug neuron (Figure \ref{fig:neuron_silencing}B).
%In addition, if we only silence the neurons with a positive phase value (that correlate with the mug handle presence), we saw an intermediate reduction, suggesting the negative phase neurons also contribute to the mug identity. 
In a sense, by selecting neurons with highest FFT amplitude, we can identify what we call \textit{mug-preferred neurons} (MPNs).

To examine the specificity of these MPNs, we tested the silencing effect on cup and car output neurons. Silencing the top MPNs has a significant impact on the logit value of the cup neuron (Figure \ref{fig:neuron_silencing}C). This is not surprising given that a mug and a cup share many common features. However, the zoomed in softmax plot shows that silencing the top $\sim$20 MPNs slightly increases the softmax value of the cup output neuron, which is consistent with the fact that most of these neurons fire when the handle is present (Figure \ref{fig:neuron_silencing}D and Supplementary Video 2). In other words, these neurons might be contributing to the difference between a mug and a cup. 

In contrast, silencing the top MPNs has almost no effect on the car output neuron (Figure \ref{fig:neuron_silencing}C). In fact, the effect of silencing MPNs is almost identical to that of silencing random neurons (dotted grey line in \ref{fig:neuron_silencing}C, see also \ref{fig:neuron_silencing}A blue line). 

These experiments showed that the MPNs contribute significantly to mug identity and much less to the identity of other categories like car. A small portion of the MPNs may be coding the handle feature to differentiate a mug from a cup.  Interestingly, although silencing one or a few neurons that are most prominent does decrease the input value to a specific output neuron, there is a significant amount of value that remains. This result confirms that object features do not seem to be represented by a single or few neurons, but rather by an ensemble of neurons.

%By performing this kind of FFT analysis on neural networks using Toybox data, we can get a powerful diagnostic tool for gaining insights into the object representations within a neural network.

% For space reason and kind of unrelated to our current version of paper, I think we have to delete the hand skeleton part --- TM

% \begin{figure}[b]
%     \centering
%     \includegraphics[width=0.7\linewidth]{hand-skeleton-v3}
%     \caption{The Toybox dataset could also serve as training data for robotic hands to learn to naturally manipulate objects. }
%     \label{fig:hand-skeleton}
% \end{figure}

%%%%%%%%%%%%%%%%%%%%%%%%%%%%%%%%%%%%%%%%%%%%%%%%%%
\section{Discussion and Future Work}
\label{sec:discussion}

In this paper, we presented the new Toybox dataset of egocentric visual object transformations.  We also provided results from two sample experiments showing how this dataset can be used to study visual learning, including (1) effects of instance diversity and view diversity on recognition performance, and (2) using a novel FFT-based method to classify hidden layer neurons according to how they represent various category- and viewpoint-dependent visual properties.
%We showed in this paper that our new Toybox dataset could complement existing dataset in studying deep convolutional neural networks. By providing structured videos showing a range of transformations, we can systematically analyze the CNN in a way that is not possible with the canonical datasets.
 
% * say something about data augmentation

%We found that increasing the object diversity and object views could enhance the performance. However, the extent to which object diversity and object views enhance the training performance is rather unexpectedly marginal after certain point. For object diversity, the training performance plateaued after around 20 objects, and more object views do not seem to further help after more than 40 views per object. We performed our experiments with transfer learning, i.e. the network is pre-trained on ImageNet. It is possible that pre-training of the network makes it less critical for large amount of objects during transfer learning. rich visual input is required to form effective early layer kernels for a deep CNN. It will be interesting to see whether one can train a deep neural network from scratch with limited number of objects but multiple views like Toybox and have a comparable performance to ImageNet.

%We showed that by applying FFT to the bottleneck neural activation, we could identify neurons that correlate with the presence or absence of specific features of the object. 

In future research, in addition to continuing the types of experiments presented here, we expect that the Toybox dataset will be valuable for studying new types of representations and learning algorithms that lend themselves to continuous image sequence inputs.
For example, in human vision, object motion is critical for segmentation, and also likely plays a role in many other aspects of object detection and recognition.  
%Most of the large-scale computer vision datasets are composed of discrete images that are randomly selected, while humans learn most vision tasks through continuous input. Toybox allows us to start addressing questions such as whether getting continuous visual input is beneficial to the performance. Besides, the ability to represent object motion and self-motion is essential for both humans and artificial intelligent agents such as self-driving cars. However, since most of the deep CNNs are built to train on random, static images, by nature, these CNNs cannot detect object motion at least from early convolutional layers. In contrast, 
%Humans have neurons both in the retina and visual cortex that prefer certain motion directions 
\cite{ohki2005functional,sabbah2017retinal}.  How motion features affect recognition performance, and how object motion might contribute to the learning phase as well (for instance, by providing a real-time version of data augmentation), are currently open research questions in the study of visual learning.
With its structured object transformations and wide selection of categories and object instances, we believe the Toybox dataset will help drive continued research advances on these and many other important questions in AI and cognitive science.
%could serve as an ideal dataset to train and test networks that could represent these desirable features of a vision system.

%%%%%%%%%%%%%%%%%%%%%%%%%%%%%%%%%%%%%%%%%%
\section{Acknowledgments}

Many thanks to Fernanda Elliot, Joshua Palmer, Soobeen Park, Joel Michelson, Aneesha Dasari, Ellis Brown, Max de Groot, Harsha Vankayalapati, and Joseph Eilbert for help in data collection. We would also like to thank early discussions influencing this research, with Linda Smith, Chen Yu, Fuxin Li, and Jim Rehg.  This research was funded in part by a Vanderbilt Discovery Grant, "New explorations in visual object recognition," and by NSF award \#1730044.

% Lastly, we think the dataset could be potentially useful for other applications. For example, the rich hand gestures (Figure \ref{fig:hand-skeleton}) could serve a training dataset for robotic hand to learn to handle objects efficiently.

% \section{Acknowledgments}
% Many thanks to Fernanda Elliot, Joshua Palmer, Soobeen Park, Joel Michelson, Aneesha Dasari, Ellis Brown, Max de Groot, Harsha Vankayalapati, and Joseph Eilbert for help in data collection. We would also like to thank early discussions influencing this research, with Linda Smith, Chen Yu, Fuxin Li, and Jim Rehg.  This research was funded in part by a Vanderbilt Discovery Grant, "New explorations in visual object recognition," and by NSF award \#1730044.
% \clearpage

% \input{Toybox_main.bbl}
% \bibliography{Toybox_main.bbl}
% \bibliographystyle{aaai}

\end{document}